# Important Molecular Descriptors Selection Using Self Tuned Reweighted Sampling Method for Prediction of Antituberculosis Activity

Doreswamy[1], Chanabasayya M. Vastrad[2]

Department of Computer Science Mangalore University, Mangalagangotri-574 199, Karnataka, India

[1]Doreswamyh@yahoo.com
[2]channu.vastrad@gmail.com

*Abstract*— In this paper, a new descriptor selection method for selecting an optimal combination of important descriptors of sulfonamide derivatives data, named self tuned reweighted sampling (STRS), is developed. descriptors are defined as the descriptors with large absolute coefficients in a multivariate linear regression model such as partial least squares(PLS). In this study , the absolute values of regression coefficients of PLS model are used as an index for evaluating the importance of each descriptor Then, based on the importance level of each descriptor, STRS sequentially selects N subsets of descriptors from N Monte Carlo (MC) sampling runs in an iterative and competitive manner. In each sampling run, a fixed ratio (e.g. 80%) of samples is first randomly selected to establish a regresson model. Next, based on the regression coefficients, a two-step procedure including rapidly decreasing function (RDF) based enforced descriptor selection and self tuned sampling (STS) based competitive descriptor selection is adopted to select the important descriptorss. After running the loops, a number of subsets of descriptors are obtained and root mean squared error of cross validation (RMSECV) of PLS models established with subsets of descriptors is computed. The subset of descriptors with the lowest RMSECV is considered as the optimal descriptor subset. The performance of the proposed algorithm is evaluated by sulfanomide derivative dataset. The results reveal an good characteristic of STRS that it can usually locate an optimal combination of some important descriptors which are interpretable to the biologically of interest. Additionally, our study shows that better prediction is obtained by STRS when compared to full descriptor set PLS modeling, Monte Carlo uninformative variable elimination (MC-UVE). Compared to the partial least squares regression models based on full descriptor set and descriptors selected by MC-UVE, the performance of STRS with PLS model was better, with higher determination coefficient for test ( $r^2$ ) of 0.8758 , and lower root mean square error of prediction of 0.1676. Based on the results, it was concluded that Sulfonamide with STRS methods seem to be a rapid and effective alternative to the classical methods for the prediction of antituberculosis activity.

*Keywords*— MC-UVE,PLS,RDF,TRS, Number of Principal factors, RMSEP ,RMSECV

## I. INTRODUCTION

Multivariate regression models have been gaining extensive applications in the analysis of multi-variant descriptors data due to their potential to extract chemically meaningful information, e.g. structure-related descriptors from the over-determined systems. But the measured bio activity data on the modern spectroscopic instrument, such as NMR and UV–V is spectra and electrochemical measurements, are usually of high colineaity, which is the common place faced by pharmaceutical chemists  To address this problem, a variety of techniques based on latent variables (LVs) have been proposed, such as principal component regression (PCR) [1,3,4] and partial least squares (PLS) [1,2,3]. Typically, the establishment of a regression model usually includes all the generated descriptors. It is obvious that such a full descriptor set model is sure to contain much redundant information, which will of course have negative influence on the prediction ability of the developed model. In addition, from the point of view of model interpretation, it is really difficult for pharmaceutcal chemists and/or chemometrists to determine which decriptors or combinations are responsible for the property of interest. It has been demonstrated that, both experimentally and theoretically, improvement of the performance of the multivariate model can be achieved by using the selected informative descriptors not the full descriptor set.

Generally, the selection criteria for  descriptors can be categorized into two groups [5]. One is based on information content of the descriptor, such as descriptors obtained from molecular graph. The other is based on the statistics related to the model's performance, e.g. RMSECV [6]

From an optimization perspective, the descriptor selection can be viewed as an optimizing process which maximizes the prediction performance of the regression model. From an optimization perspective, the descriptor selection can be viewed as an optimizing process which maximizes the prediction performance of the regression model. Thus, it is natural to employ the optimization algorithm, which tries to seek a good combination of descriptors, to implement descriptor selection using the criteria mentioned above as the objection function. Genetic algorithm (GA) [7-11],Entropy based variable selection[12-14], Filter Type Methods [15-17], Simulated Anealing [18-20] have been applied to select the optimal subset of descriptors. All these studies suggest that better prediction can be obtained using the selected descriptors rather than the full set of descriptors, which is an indication of the importance of descriptor selection. But one should know that this kind of methods based on optimization methods is usually computationally intensive and sensible to the initialized solution.

Partial Least Squares is intended as a full set of descriptor regression method, but this often necessitates pre-treatment of the data to reduce the number of non-informative variables(descriptors) to an acceptable level prior to bilinear modelling. Elimination of uninformative descriptors can predigest regression modelling and improve prediction results in terms of accuracy and robustness. Better quantitative regression models may be obtained by selecting characteristic descriptors including sample-specific or component-specific information instead of the full set of descriptor. Besides, a series of more direct methods have been proposed to conduct descriptor selection, such as iterative partial least squares (iPLS) [21], uninformative variable elimination (UVE) [22-23], Monte Carlo based UVE (MC-UVE)[24,25] and so on. These algorithms to increase the predictive ability of the standard PLS method where descriptors (or independent variables) which can not contribute to the model construction very much are eliminated.

The objective of this paper , we present a new strategy, termed self tuned reweighted sampling , which has the potential to select an optimal combination of the descriptors existing in the full set of descriptors coupled with partial least squares regression and to compare the prediction results based on selected descriptors and full descriptor. To compare and determine the effective descriptors selected by MC-UVE for prediction of antituberculosis activity.

## II. MATERIALS AND ALGORITHMS

### A. *The Data Set*

The molecular descriptors of 100 Sulfonamide derivative [26,27] based H37Rv inhibitors analysed. These molecular descriptors are generated using Padel-Descriptor tool [28]. The dataset covers a diverse set of molecular descriptors with a wide range of inhibitory activities against H37Rv. The pIC50(observed biological activity) values range from 4.06 to 8. The dataset can be arranged in data matrix. This data matrix X contains m samples(molecule structures) in rows and p descriptors in columns. Vector y with order m×1 denotes the measured activity of interest i.e pIC50. When modeling, both X and y are mean-centered.

### B. *Monte Carlo Uninformative Variable Elimination*

MC-UVE is a frequently used variable(descriptor) selection method which combined Monte Carlo strategy with uninformative variable elimination method. The MC-UVE method builds a large number of models with randomly selected training samples at first, and then, each variable is evaluated with a stability of the corresponding coefficients in these models. Variables with poor stability are known as uninformative variable and eliminated.

### C. *Self tuned reweighted sampling*

The proposed method works in four successive steps: (1) Monte Carlo for model sampling. (2) Employ Rapidly decreasing function(RDF) to perform enforced descriptors selection. (3) Adopt tuned reweighted sampling(TRS to realize a competitive selection of descriptors and (4) cross validation [29–31] is utilized to evaluate the subset. STRS will be discussed in great detail in the following sections. The STRS algorithm is implemented in MATLAB.

### D. *Monte Carlo for model sampling*

Like uninformative variable elimination [22,23], in each sampling run of STRS, a PLS model is built using the randomly selected samples (usually 80% of the training sample) not all the samples in the training set. From the point of view of sampling, this process can be regarded as sampling in the model space combined with Monte Carlo strategy. We are intended to select the descriptors which are of high adaptability regardless of the variation of training samples.

### E. *PLS and weights of descriptors*

PLS is a widely used procedure for modelling the linear relationship between X and y based on latent variables (LVs). Suppose that the scores matrix is denoted by T, which is a linear combination of X with W as combination coefficients, and c is the regression coefficient vector of y against T by least squares. Thus we have the following formula:

$$T = XW \quad \text{-----------------------------------(1)}$$

$$y = Tc + e = XWc + e = Xb + e \quad \text{------(2)}$$

where e is the prediction error and $b = Wc = [b_1, b_2, \ldots b_3]^T$ is the p-dimensional coefficient vector. The absolute value of the ith element in b, denoted $|b_i|$ (1≤i≤p) reflects the ith descriptors's contribution to y. Thus, it is natural to say that the larger $|b_i|$ is, the more important the ith descriptor is. For evaluating the importance of each descriptor, we define a normalized weight as:

$$w_i = \frac{|b_i|}{\sum_{i=1}^{p} |b_i|}, \quad i = 1, 2, 3, \ldots p \quad \text{---(3)}$$

Additional attention should be paid to that the weights of the eliminated descriptors by STRS are set to zero manually so that the weight vector w is always p-dimensional.

### F. Rapidly decreasing function

Suppose the full spectrum contains p descriptors and N sampling runs are performed in STRS. As mentioned before, the descriptor selection in STRS consists of two steps. In the first step, RDF is utilized to remove the descriptors which are of relatively small absolute regression coefficients by force. In the ith sampling run, the ratio of descriptors to be kept is computed using an RDF defined as:

$$r_i = ae^{-ki} \quad \text{--------------------(4)}$$

where a and k are two constants determined by the following two conditions: (I) in the first sampling run, all the p descriptors are taken for modeling which means that $r_1 = 1$, (II) in the Nth sampling run, only two descriptors are reserved such that we have $r_N = 2/p$. With the two conditions, a and k can be calculated as:

$$a = \left(\frac{p}{N}\right)^{1/(N-1)} \quad \text{------------------------(5)}$$

$$k = \frac{\ln(p/2)}{N-1} \quad \text{--------------------------(6)}$$

where ln denotes the natural logarithm

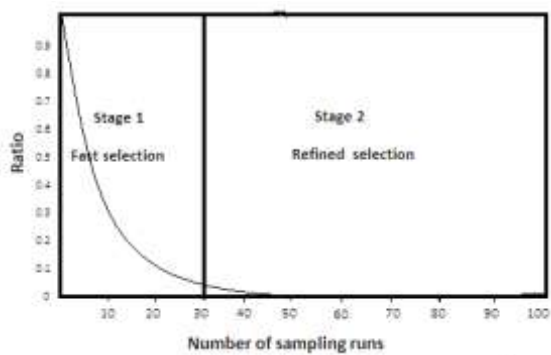

Fig. 1 Graphical illustration of the rapidly decreasing function. In the first stage, the number of the descriptors is reduced fast while in the second stage, it decreases very slowly which realizes a refined selection.

Fig. 1 illustrates an example of RDF. As can be seen clearly, the process of descriptor reduction can be roughly divided into two stages. In the first stage, descriptors are eliminated rapidly which performs a 'fast selection', whereas in the second stage, descriptors are reduced in a very gentle manner, which is instead called a 'refined selection' stage in our study. Therefore, descriptors of little or no information in a full set of descriptor can be removed in a stepwise and efficient way because of the advantage of RDF. That is the reason why we choose RDF. Its advantage will be demonstrated by our experiments in the following sections.

### G. Tuned reweighted sampling

Following RDF-based enforced descriptors reduction, tuned reweighted sampling (TRS) is employed in STRS to further eliminate descriptors in a competitive way. Fig. 2 illustrates the meaning of tuned reweighted sampling. Assume that we have five weighted descriptors which will be subjected to five random weighted sampling experiments with replacement. In Case 1, each descriptor has an equal weight 0.20 indicating that they can be sampled with an equal probability. The ideal result is that each descriptor is sampled one time. Case 2 shows descriptors 1 and 2 have the largest weight 0.30 while descriptors 4 and 5 are of the smallest weights 0.10. Thus, descriptors 1 and 2 are sampled twice, while descriptor 3 once. descriptors 4 and 5 are not sampled by TRS and hence eliminated. Similar to Case 2, Case 3 demonstrates that only descriptors 1 and 3 are sampled in the five weighted sampling experiments due to their dominant weights, while descriptors 2, 4 and 5 are much less competitive and hence out of play because of their relatively weak weights.

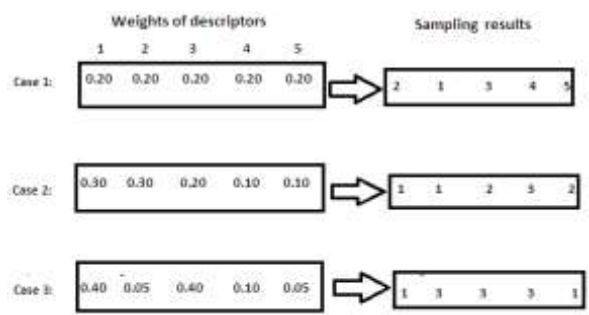

Fig. 2 Illustration of tuned reweighted sampling technique using five variables in three cases as an example. The descriptors with larger weights will be selected with higher frequency.

*Doreswamy et al.*

### H. Algorithmic description of STRS

The STRS algorithm selects N subsets of descriptors by N sampling runs in an iterative manner and finally chooses the subset with the lowest RMSECV value as the optimal subset. In each sampling run, STRS works in four successive steps including Monte Carlo model sampling, enforced descriptors reduction by RDF, competitive descriptors reduction by TRS and RMSECV calculation for each subset. Of these, RDF-based descriptors reduction in combination with competing descriptor reduction by TRS is a two-step procedure for descriptor selection. The STRS procedure steps are explained here.

- Input descriptor data set X and its Activity data(pC50) y.
- While i <=N sampling runs if it fails go to step 8.
- Randomly choose K samples ($X^k, y^k$) using $D_{selectold}$ to build PLS model.
- Record the absolute regression coefficients $b_j$ then $W_j = b_j/sum(b_j)$.
- Compute the ratio of descriptors to be kept using $r_i = ae^{-ki}$.
- Pick subset of descriptors from the retained p x $r_i$ descriptors using tuned reweighted sampling
- method, denoted by $D_{selectnew}$.
- Compute RMSECV using D. Then, $D_{selectold} = D_{selectnew}$. go to step 2
- After N sampling runs, STRS obtains N subsets, STRS obtains N subsets of descriptors and
- corresponding N RMSECV values
- Finally, choose the subset with the lowest RMSECV as optimal subset of descriptors.

### I. Model Performance Criteria

The statistics used for estimating the performance of the regression models included coefficient of determination for prediction $r^2$ and root mean square error of prediction (RMSEP).

$$r^2 = 1 - \frac{\sum_{i=1}^{n}(y_i - y_{i(pred)})^2}{\sum_{i=1}^{n}(y_i - y_{mean})^2} \quad \text{-----------------------(5)}$$

$$RMSEP = \sqrt{\frac{\sum_{i=1}^{n}(y_i - y_{i(pred)})^2}{n}} \quad \text{-----------------------(6)}$$

### J. RESULTS AND DISCUSSION

#### A. Influence of number of MC sampling runs

In order to investigate the influence of the number of Monte Carlo sampling runs on STRS' performance, we have considered the following four cases: the number is set to 50, 100, 200 and 500. For each case for the dataset, 50 replicate running of STRS is executed and RMSECV values are recorded. The resulted statistical box-plots are shown in Fig. 3. It can be found that the number of Monte Carlo sampling runs does not have significant influence on the performance of STRS.

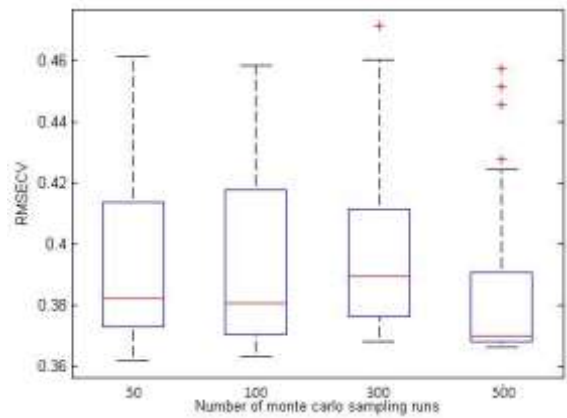

Fig. 3. The box-plots for the dataset with the number of Monte Carlo sampling runs of STRS set to 50, 100, 200 and 500, respectively.

#### B. Determination of the principal factor number (nLV) for PLS modelling

The number of principal factor (nLV) of PLS is an important parameter in the modelling. Therefore, in this work, The parameter is determined with the root mean squared error of prediction (RMSEP) of the assessing set and the RMSEP of the training set in cross-validation (denoted by RMSECV). Fig. 4 show the variation of RMSEP and RMSECV with the principal factor number of the three methods, i.e., STRS-PLS, MCUVE-PLS and PLS methods for the data set. From the figure, it was clear that both RMSEP and RMSECV have a descending trend with the increase of the principal factor number, but the trend slowed down after nLV > 10. Therefore, Monte Carlo cross-validation with F-test was used for confirming the suitable principal factor number, and the results show that 10–15 can be used. In order to make the model as less as complex and use an identical parameter in the three models, nLV= 10 was used further calculations.

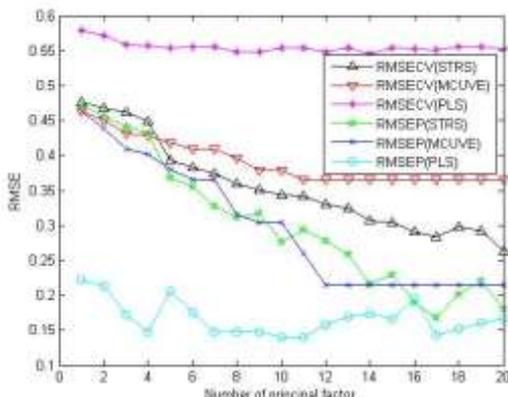

Fig. 4 Variation of RMSEP and RMSECV with the number of factors by STRS, MC-UVE and PLS methods for Sulfonamide data set

### C. Comparative Analysis for the Sulfonamide dataset STRS, MC-UVE and PLS methods

This Sulfonamide descriptor data is employed to specially address the situation that much better prediction results can only be obtained by combination of some important descriptors which are interpretable to the biological activity of interest. 10-fold cross validation is used in this study to explore its predictive performance. Also, we compared STRS to MC-UVE, aiming only at demonstrating that STRS is indeed an alternative and efficient procedure for uninformative variable elimination not that which method is better.

TABLE 1: THE RMSECV RESULTS ON THE SULFONAMIDE DATASET FOR PLS, MC-UVE, STRS. THE NLVS AND NVAR DENOTES THE NUMBER OF LATENT VARIABLE S AND SELECTED VARIABLES, RESPECTIVELY

| Methods | RMSECV | nLVs | nVAR |
|---------|--------|------|------|
| PLS     | 0.5451 | 2    | 729  |
| MC-UVE  | 0.3654 | 2    | 118  |
| STRS    | 0.2635 | 15   | 29   |

This data is first cantered for each descriptor to have zero mean and unit variance before modelling. By 10-fold cross validation, the optimal number of latent variables of PLS model is 2. For MC-UVE, the number of Monte Carlo iterations is set to 500, and in each iteration 80% samples from this data are randomly chosen to build a PLS regression model using two latent variables. The regression coefficients for each descriptor are recorded in a vector. After 500 iterations, a coefficient matrix is obtained based on which a reliability index can be calculated for each descriptor. Then, all the descriptors are ranked in accordance with their reliability index. As known, cross validation is an effective and widely used technique for model/descriptor selection. Thus in our study, the number of descriptors to be selected is determined by 10-fold cross validation technique. Also the maximal number of selected descriptors is set to 365. With these settings, we run MC-UVE to eliminate the uninformative descriptors while simultaneously estimate its predictive performance. Further, it is noteworthy that only one running of MC-UVE is not sufficient due to the variation caused by Monte Carlo strategy. One solution for this problem is to repeat it for many times. Therefore, MC-UVE is repeated 500 times in this case, which can help to get a deeper understanding of its behavior. For STRS, the number of MC sampling runs is set to 100. STRS is also rerun for 500 times and the results are recorded for further analysis.

Table 1 shows the results of MC-UVE and STRS on Sulfonamide descriptors data, together with the results based on the full descriptor set and only the informative variables. The RMSECV value using all the 729 descriptors is 0.5451. By contrast, not only the RMSECV (0.3654) but also the number of latent variables is not changed significantly when the model only selects the subset of the 365 informative descriptors. This phenomenon experimentally proves the necessity to perform descriptor selection or removing the uninformative descriptors before building a regression model.

MC-UVE and STRS are applied in order to demonstrate whether better prediction can be obtained by selecting the reliable descriptors (MC-UVE) or important descriptors (STRS). From Table 1, one can find that STRS got much better prediction results, i.e. 0.2635 compared to 0.3654, which indicates that the stability of STRS still needs improving although it can pick out descriptors leading to a model with good generalization performance. Interestingly, the number of the selected descriptors by STRS is relatively small i.e. 29, which is one reason why we call them important descriptors. Moreover, only one uninformative descriptor is selected one time by STRS, which proves that it has the potential to eliminate uninformative descriptor as MC-UVE does.

*Doreswamy et al.*

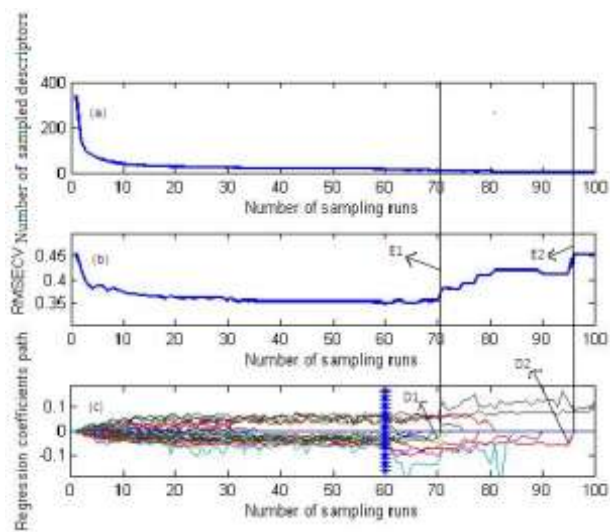

Fig. 5 The changing trend of the number of sampled descriptors (plot a), 10-fold RMSECV values (plot b) and regression coefficients of each descriptors (plot c) with the increasing of sampling runs. The line (marked by asterisk) denotes the optimal point where 10-fold RMSECV value achieve the lowest.

Fig. 5 shows the changing trend of the number of sampled descriptors (plot a), 10-fold RMSECV values (plot b) and the regression coefficient path of each descriptor (plot c) with the increasing of sampling runs from one STRS running. As expected, the number of sampled descriptors decreases fast at the first stage of RDF and then very slowly at the second stage of RDF, which demonstrated that the proposed two phase selection, i.e. fast selection and refined selection, are indeed realized in STRS. The RMSECV values first descend quickly from sampling runs 1–35 which should be the elimination of uninformative descriptors, then changes in a gentle way from sampling runs 36–58 corresponding to the phase that the sampled descriptors do not change obviously, and finally increase fast because of the loss of information caused by eliminating some important descriptors from the optimal subset (denoted by stars symbol).

Also noteworthy is the coefficient path of each descriptor shown in plot c. Each line in plot c records the coefficients at different sampling runs for each descriptor. Thus, a subset of descriptors together with the regression coefficients can be extracted from each sampling run. The best subset with the lowest RMSECV value is marked by the vertical line denoted by asterisk. More interestingly, the RMSECV value jumps up to a higher stage at the sampling point (denoted solid line:E1), because the coefficient of one descriptor (denoted by D1) drops to zero just at the same time. The sold line marked by E2 is also the case when the coefficient of another descriptor denoted by D2 drops to zero. Such observations demonstrate the existence of important descriptors without which the model's performance would be reduced dramatically. That is why they are called important descriptors. In general this study indicates that STRS is a promising method for descriptor selection.

The training set with 75 samples was used to build all three model and the performance is also evaluated by 25 samples in test set. First PLS algorithm was applied based on full set of descriptor (i.e 729 descriptors). Fig 6 is the scatter plot of the model, which shows a correlation between observed value and antituberculosis activity prediction in the training and test set. And the values of RMSEP = 0.4711 and $r^2$ = 0.7575 are obtained.

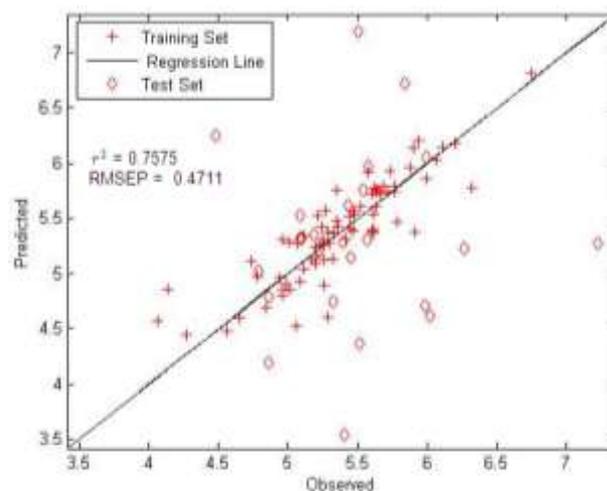

Fig. 6 correlation between observed and predicted values for training set and test set for full descriptor set

By using the twenty nine selected descriptors from STRS descriptor selecton method with a partial least squares(PLS) regression model. The prediction results are shown in Fig. 7.

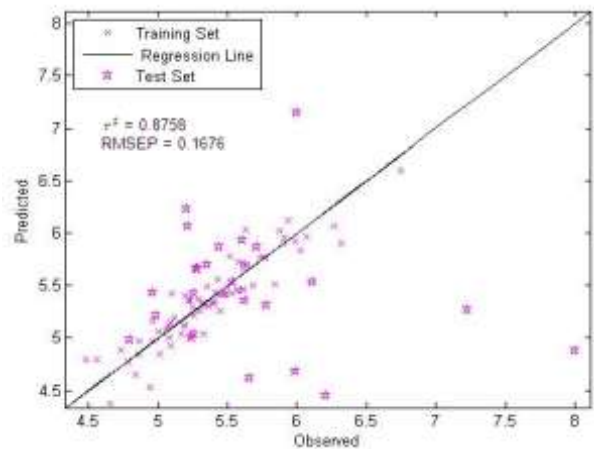

Fig .7 correlation between observed and predicted values for training set and test set after self tuned reweighted sampling descriptor selection.

As can be seen, the value of RMSEP = 0.1676 is smaller than the one obtain on full set of descriptor, and the value of $r^2$ = 0.8758 is larger than the one obtained on full set of descriptors. This reveals that descriptor selection is helpful for prediction of antituberculosis activity.

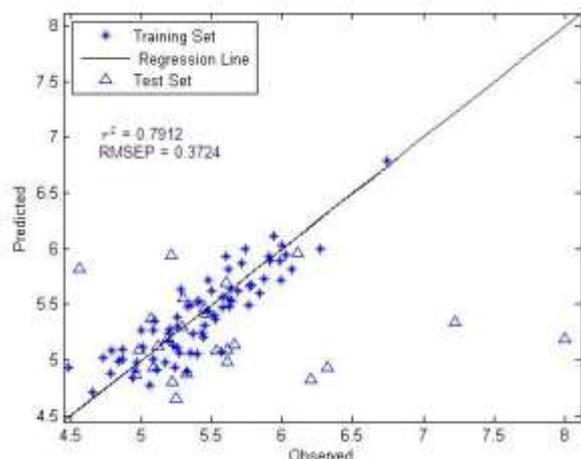

Fig .8 correlation between observed and predicted values for training set and test set after MC-UVE descriptor selection

By using the 118 selected descriptors from MC-UVE method with a partial least squares(PLS) regression. The prediction results are shown in Fig. 8. As can be seen, the value of RMSEP=0.3724 is smaller than the one obtained on full set of descriptor but larger than one obtained on the descriptors selected by STRS. It should be noted that both STRS and MC-UVE adopt Monte Carlo strategy to perform descriptor selection. Therefore, the selected descriptors are not exactly the same for each run. It is necessary to run the programs many times to obtain statistically stable results.

The overall results indicate that Sulfonamide derivatives data combined with STRS is successfully applied for the determination of antituberculosis activity. Moreover, the results demonstrated that STRS is a powerful way for the selection of effective descriptors for predictive analysis. This would be helpful for us to understand the correlation between the descriptor and antituberculosis activity.

### D. CONCLUSIONS

This paper presents a new method for important descriptor selection using self tuned reweighted sampling technique coupled with PLS. Based on the importance level of each descriptors ,STRS sequentially selects N subsets of descriptors from N sampling run. In each sampling run, the number of descriptors tobe selected by STRS is controlled by the proposed rapidly decreasing function and further by tuned reweighted sampling. In an efficient and competitive way, STRS finally selects a combination of important descriptors which is of great importence. This method is applied to sulfonamide dataset , it is demonstrated that STRS is a promising procedure to eliminate the uninformative descriptors and/or conduct descriptor selection for building a high performance regression model.

Descriptor selection is necessary for prediction of antituberculosis activity. Proper descriptor selection can reduce the complexity of the regression model and improve the prediction accuracy. Compared to MC-UVE, STRS is a powerful method for the selection of important descriptors for this application. Summarizing, the Sulfonamide dataset coupled with STRS methods seem to be a rapid and effective alternative to the classical methods for the prediction of antitubercular activity.

#### ACKNOWLEDGEMENT

We gratefully thank to the Department of Computer Science Mangalore University, Mangalore India for technical support of this research.